\documentclass[conference]{IEEEtran}
\IEEEoverridecommandlockouts
\usepackage{cite}
\usepackage{amsmath,amssymb,amsfonts}
\usepackage{textcomp}
\usepackage[table,xcdraw]{xcolor}
\usepackage{marvosym}
\usepackage{subfig}
\usepackage{graphicx,color}
\usepackage{algorithm}
\usepackage{algorithmic}
\usepackage{bm}
\usepackage{soul}
\usepackage{colortbl}
\usepackage{multirow}
\usepackage{multiobjective}
\usepackage[numbers]{natbib}

\newcommand\algName{mBOA}
\newcommand\algNameLong{Multi-objective Bayesian Optimization Algorithm}

\newcounter{ALC@tempcntr}
\newcommand{\LCOMMENT}[1]{%
    \setcounter{ALC@tempcntr}{\arabic{ALC@rem}}
    \setcounter{ALC@rem}{1}
    \item \{#1\}
    \setcounter{ALC@rem}{\arabic{ALC@tempcntr}}
}%
\def\BibTeX{{\rm B\kern-.05em{\sc i\kern-.025em b}\kern-.08em
    T\kern-.1667em\lower.7ex\hbox{E}\kern-.125emX}}

\begin{document}

\title{On the performance of multi-objective estimation of distribution algorithms for combinatorial problems}


\author{\IEEEauthorblockN{Marcella S. R. Martins}
\IEEEauthorblockA{\textit{Federal University  of Technology - Paran\'{a}} \\
Email: marcella@utfpr.edu.br}
\and
\IEEEauthorblockN{Mohamed El Yafrani}
\IEEEauthorblockA{\textit{Mohammed V University of Rabat} \\
Email: m.elyafrani@gmail.com}
\and
\IEEEauthorblockN{Roberto Santana}
\IEEEauthorblockA{\textit{University of the Basque Country} \\
Email: roberto.santana@ehu.es}
\and
\IEEEauthorblockN{Myriam  Delgado}
\IEEEauthorblockA{\textit{Federal University  of Technology - Paran\'{a}} \\
Email: myriamdelg@utfpr.edu.br}
\and
\IEEEauthorblockN{Ricardo L\"{u}ders}
\IEEEauthorblockA{\textit{Federal University  of Technology - Paran\'{a}} \\
Email: luders@utfpr.edu.br}
\and
\IEEEauthorblockN{Bela\"{i}d Ahiod}
\IEEEauthorblockA{\textit{Mohammed V University of Rabat} \\
Email: bahiod@hotmail.com}
}


\maketitle

\begin{abstract}
Fitness landscape analysis investigates features with a high influence on the performance of optimization
algorithms, aiming to take advantage of the addressed problem characteristics.
In this work, a fitness landscape analysis using problem features is performed for a \algNameLong~(\algName) on instances of MNK-landscape problem for 2, 3, 5 and 8 objectives. We also compare the results of \algName~with those provided by NSGA-III through the analysis of their estimated runtime necessary to identify an 
approximation of the Pareto front. Moreover, in order to scrutinize the probabilistic graphic model obtained by \algName, the Pareto front is examined according to a probabilistic view.
The fitness landscape study shows that \algName~is moderately or loosely influenced by some problem features, according to a simple and a multiple linear regression model, which is being proposed to predict the algorithms performance in terms of the estimated runtime. Besides, we conclude that the analysis of the probabilistic graphic model produced at the end of evolution can be useful to understand the convergence and diversity performances of the proposed approach.
\end{abstract}


\section{Introduction}

Population-based metaheuristics and Evolutionary Algorithms (EA) have been applied to solve multi-objective optimization problems, mainly due to their ability to discover multiple solutions in parallel and to handle the complex features of such problems~\cite{Coello:1999}. Besides, probabilistic modeling can also be aggregated to capture and exploit the potential regularities that arise in the promising solutions, which is the basis of an Estimation of Distribution Algorithm (EDA)~\cite{muhlenbein:96}.

The main idea of EDAs is to extract and represent, using a probabilistic model, the regularities shared by a subset of high-valued problem solutions. New solutions are then sampled from the probabilistic model to guide the search toward areas where optimal solutions are more likely to be found. Normally, Multi-objective Estimation of Distribution Algorithm (MOEDA)~\cite{kasrchenas:14} integrates both model building and sampling techniques into evolutionary multi-objective optimizers using special selection schemes. Probabilistic graphical models, which combine graph and probability theory, have been broadly used to improve EDAs and MOEDAs performance~\cite{larranaga:12}. As reported in \cite{mar:16}, most of MOEDAs developed to deal with combinatorial multi-objective optimization problems (MOPs) adopt Bayesian Networks as their PGM.

One of the main challenges in multi-objective optimization is to find the Pareto optimal set, or an approximation of it. The Pareto set plays a central role in the search space structure. The definition and analysis of fitness landscape for MOPs can help to understand the geometry of a combinatorial MOP, for example, and to explain the ability multi-objective metaheuristics to obtain an approaximation of the Pareto set.

The fitness landscape of a problem instance is the topological structure over which a search is being executed~\cite{watson2010introduction}, defined by the solution candidates, their neighborhood structure and the fitness of the solution candidates.

By addressing the relative importance of features in explaining the metaheuristic performance variance, fitness landscape analysis (FLA) allows one to investigate which features of a combinatorial MOP have the highest influence on the metaheuristic performance. Through the study of these features, one should be able to design metaheuristic algorithms to take advantage of the multi-objective optimization characteristics of a given problem~\cite{garrett2007multiobjective}.

Several works have addressed combinatorial MOPs from an FLA perspective.
\citet{borges2002study} presented a study of global convexity for a multi-objective travelling salesman problem. The authors investigated features concerning instances solved by scalarizing algorithms (which use weights to aggregate multiple objectives), like the distribution of local optima, the differences observed between solutions and points generated with different weights, as well the stability of the best local optima for small weight variations. The exact global optima was also generated and the results confirmed the existence of global convexity that might be useful in the multi-objective optimization context.

\citet{garrett2007multiobjective} provided a high-level overview of multi-objective search space and FLA. They investigated some features like the distribution of local and Pareto optima, fitness distance correlation, ruggedness and random walk on the multi-objective generalized assignment problem.

This paper addresses a combinatorial MOP - multi-objective NK-landscape (MNK) model which has been recently explored in other works in the literature~\cite{aguirre2007working,santana2015multi,liefooghe2015feature}. 
In particular, EDAs that use different types of probabilistic models, including Bayesian Networks, have already been applied to the mono-objective NK problem~\cite{pelikan2008analysis,pelikan2009performance,liaw2013effect}.

The objective here is to extend those works to multi and many objective optimization (when the number of objectives is higher than $3$) conducting an FLA to investigate the impact of instance
features on the search performance of a MOEDA based on Bayesian Network (BN) as the PGM structure, called \algNameLong~(\algName)~\cite{khan:02}. The algorithm is compared with NSGA-III~\cite{deb:14}, a state-of-the-art algorithm applied to solve multi and many-objective optimization problems (MaOPs).

Our main contribution is the use of FLA to explore the relationship between performance measures and relevant problem features for mBOA. As far as we know, this is the first work to consider FLA for MOEDAs. 
In addition, our work also includes an investigation, from a probabilistic point of view, of the Pareto-front using the probability mass function provided by mBOA at the end of evolution. 

This paper has intersections with other previously published works. It is linked to the work presented by~\citet{liefooghe2015feature} in which some features of the MNK model instances are investigated. It is also related to the work developed by ~\citet{echegoyen2012toward} which considers a quantitative analysis to compare the probabilities of sampling the optimum and the most probable solution between successful and failed trials aiming to understand the behaviour of EDAs based on BNs for mono-objective problems. However, differently from~\cite{liefooghe2015feature}, we are investigating an EDA approach, and extending the analysis to 8 objectives. Also, in contrast to the research presented in~\cite{echegoyen2012toward}, which investigates mono-objective optimization, our work considers a set of non-dominated solutions. Therefore, the probabilities might be calculated according to the distribution of these solutions over the true Pareto front.


\section{Background} \label{sec:background}

\subsection{The MNK-landscape model}

The \emph{NK fitness landscapes} is a family of problems proposed by~\cite{kauffman1993origins} in order to explore the way in which the neighborhood structure and the strength of the interactions between neighboring variables are linked to the ruggedness of search spaces. For the given parameters, the problem consists in finding the global maximum of the function~\cite{santana2015evolving}.

Let $ {\bf{X}}=(X_1,\ldots ,X_N)$ denote a vector of discrete variables and ${\bf{x}}=(x_1,\ldots ,x_N)$ an assignment to the variables.

An NK fitness landscape is defined by the following components~\cite{pelikan2009performance}:

\begin{itemize}
 \item Number of variables, $N$.
 \item Number of neighbors per variable (ruggedness), $K$.
 \item A set of neighbors, $\Pi(X_n) \in {\bf{X}}$, for $X_n$, $n \in \{1,\dots, N\}$ where $\Pi(X_n)$ contains $K$ neighbors.
 \item A subfunction $f_n$ defining a real value for each combination of values of $X_n$ and $\Pi(X_n)$, $n \in \{1,\dots, N\}$.
\end{itemize}

Both the subfunction $f_n$ for each variable $X_n$ and the neighborhood structure $\Pi(X_n)$
are randomly set~\cite{pelikan2009performance}).

The mono-objective function $z_{NK}$ to be maximized is defined as:
\begin{equation}
  z_{NK}({\bf{x}}) = \sum_{n=1}^{N} f_n(x_n ,\Pi(x_n)). \label{eq:FNK}
\end{equation}



MNK-landscape~\cite{aguirre2007working} is a multi-objective combinatorial optimization problem with $2$ or more objectives, 
where each objective function is determined by a different instance of the NK-landscape model
${\bf{z(\mathbf{x})}}= (z_1(\mathbf{x}),z_2(\mathbf{x}),\dots,z_M(\mathbf{x})): \mathcal{B}^N \rightarrow \mathcal{R}^M$, over the same binary string $\mathbf{x}$, where $N$ is the number of variables, $M$ is the number of objectives, $z_m(\mathbf{x})$ is the $m$-ith objective function, and $\mathcal{B}=\{0,1\}$. ${\bf{K}}=\{K_1,\dots,K_M\}$ is a set of integers where $K_m$ is the size of the neighborhood in the $m$-th landscape.

The MNK-landscape problem can be formulated as follows:

\begin{equation}
\begin{array}{l}
 \max\limits_{\mathbf{x}}\,\mathbf{z}(\mathbf{x})=(z_1(\mathbf{x}),...,z_M(\mathbf{x})) \\
 \mbox{subject to} \, \mathbf{x}\in\{0,1\}^N,\;\\\\
     \mbox{with} \,\\
     \mbox{} \, z_m({\bf{x}}) = \frac{1}{N} \sum_{n=1}^{N} f_{m,n}(x_n ,\Pi_{m,n}(x_n)),  \\
     \mbox{ } m \in \{1,...M\},  \\
     \mbox{ } n \in \{1,...N\},  \\
\end{array}
\end{equation}
where the fitness contribution $f_{m,n}$ of variable $x_n$  is a real number in $[0,1]$ drawn from a uniform distribution.

\subsection{Fitness Landscape}

Fitness landscapes illustrate the association between search and fitness space~\cite{watson2010introduction}. Given a specific landscape structure, an evolutionary algorithm can be seen as a strategy for navigating this structure in the search for optimal solutions. Therefore, fitness landscapes have been applied to investigate the dynamics of evolutionary and heuristic search algorithms for optimization and design problems~\cite{liefooghe2013makes}. In addition, the study of fitness landscapes can help predicting the performance of those algorithms.

Cost models have been used to make specific predictions regarding the behavior of evolutionary algorithms  
identifying the fitness lansdscape features that make a problem more or less difficult to solve~\cite{watson2010introduction}. 
Cost models are expressed as linear or multiple regression models of features and search cost.

There are several problem features that define the structure of fitness landscapes and can influence the difficulty level during the search. In this paper we are interested in the MNK-landscape problem, which has been explored by~\cite{aguirre2007working,santana2015multi}, for example,
in order to study algorithm's behavior with respect to a set of relevant problem features. 
The features examined in our work are based on~\cite{liefooghe2015feature} and will be presented in Table \ref{tab:feature-stats} of Section \ref{sec:experiments}.

\subsection{Bayesian Network}

A Bayesian Network (BN) is a probabilistic model that consists of a directed acyclic graph (DAG)
whose nodes represent variables, and whose edges express the probabilistic dependency between them~\cite{pearl:88}.

Let us assume $\mathbf{Y}=(Y_1,...,Y_M)$ as a vector of random variables, where $y_m$ is a value of the $m$-th
component ($Y_m$) of the vector $\mathbf{Y}$.  The set of conditional
dependencies of all variables in $\mathbf{Y}$ is described by the DAG structure $B$.
$\mathbf{Pa}^B_m$ represents the set of parents of the variable $Y_m$ given by $B$, 
and the set of local parameters $\Theta$ contains, for each variable, the conditional probability distribution of its values given different value settings for its parents, according to structure $B$.

Therefore, a Bayesian Network encodes a factorization for the probability mass function (pmf) as follows:

\begin{equation}
\label{eq:rb}
p(\mathbf{y})=p(y_1,y_2,...,y_M)=\prod_{m=1}^{M}p(y_m|\mathbf{pa}_{m}^B)
\end{equation}

We can assume, in discrete domains, that $Y_m$ has $s_m$ possible values, $y^1_m,..., y^{s_m}_m$, therefore the
particular conditional probability, $p(y_m^k|\mathbf{pa}^{j,B}_m)$ can be defined as:

\begin{equation}
\label{eq:disc_dist}
p(y_m^k|\mathbf{pa}^{j,B}_m)=\theta_{y_m^k|\mathbf{pa}^{j,B}_m}=\theta_{mjk}
\end{equation}
where $\mathbf{pa}_m^{j,B} \in \{ \mathbf{pa}_m^{1,B},...,\mathbf{pa}_m^{t_m,B}\}$  denotes a particular combination of values for $\mathbf{Pa}^B_m$ and
$t_m$ is the total number of different possible instantiations of the parent variables of $Y_m$ given by $t_m=\prod_{Y_v\in \mathbf{Pa}_m^B} s_v$, where $s_v$ is the total of possible values (states) that $Y_v$ can assume.
The parameter $\theta_{mjk}$ represents the conditional probability that variable $Y_m$ takes
its $k-$th value ($y_m^k$), knowing that its parent variables have taken their $j$-th combination of values ($\mathbf{pa}^{j,B}_m$).

The parameters $\theta_{mjk}$ can be estimated based on the current data $D$ with $N$ observations (instantiations) of $\mathbf{Y}$ \footnote{In Section \ref{sec:edas} $D$ is the population set $Pop$ with $P$ observations.} using Bayesian Estimate, where the expected value $E(\theta_{mjk}|\mathbf{N}_{mj}, B)$ of $\theta_{mjk}$ is given by Equation \ref{eq:est_bayes}:

\begin{equation}
\label{eq:est_bayes}
\hat{\theta}_{mjk}=(1 + N_{mjk})/(s_m + N_{mj})
\end{equation}
where $N_{mjk}$ is the number of observations in $D$ for which $Y_m$ assumes the $k$-th value given the $j$-th combination of values from its parents and $\mathbf{N}_{mj}=\{N_{mj1},...,N_{mjs_m}\}$.

%
%
%

To learn the BN parameters and the structure, the Bayesian Estimate and the K2~\cite{cooper:92} algorithm is used, respectively. K2 is a  greedy local based procedure that optimizes a score that measures the quality of the BN structure.

A BN is used as the PGM for \algName, whose performance is examined here using FLA concepts associated with the exploration of the final PGM model.

\section{Multi-objective Estimation of Distribution Algorithm}
\label{sec:edas}

In this paper, we consider a MOEDA called mBOA~\cite{khan:02} based on Bayesian Network as the probabilistic model and Pareto dominance as the selection scheme.

\subsection{The \algNameLong}

The framework for the MOEDA considered here is presented in Algorithm~\ref{alg:mobeda_framework}.

\begin{algorithm}[h]
\scriptsize
\caption{MOEDA framework} \label{alg:mobeda_framework}
\begin{algorithmic}[1]

\REQUIRE {$\textit{Instance}$: problem instance\\
\hspace{7mm}$P$: population size\\
\hspace{7mm}$P_\textit{PGM}$: number of solutions selected to support the probabilistic\\
\hspace{16.5mm} model estimation\\
\hspace{7mm}$P_\textit{smp}$, number of solutions sampled from the probabilistic model\\
\hspace{7mm}$\textit{T}_\textit{max}$: maximum number of evaluations\\
}
\ENSURE{$\textit{Pop}_\mathrm{ND}$: the set of non-dominated solutions}

\LCOMMENT{Initialization}

\STATE $I \gets $ \textit{LoadInstance}($\textit{Instance}$)
\STATE $\textit{Pop}^1 \gets $ \textit{RandomGenerate}$(P,I)$ \COMMENT{initial population}
\STATE $g \gets 1$

\LCOMMENT{Main loop}
\REPEAT
    \FOR{each solution $\mathbf{x} \in Pop^g$}
      \STATE $\textit{fitness}(\mathbf{x})\gets $\textit{EvaluateFitness}($\mathbf{x},I$)
    \ENDFOR

    \LCOMMENT{Non-dominated Sorting}
    \LCOMMENT{Defines $Tot_F$ Pareto fronts from the best ($i=1$) to the worst, and assigns a crowding distance}
    \STATE $F_1...F_{Tot_F}\gets $ \textit{ParetoDominance}($\textit{Pop}^g$);
    \STATE{$\textit{Pop}^g =(F_1\cup ...\cup F_{Tot_F})$;}

    \LCOMMENT{EDA: learning the probabilistic model}
    \STATE $\textit{Pop}_\mathrm{PGM}^g\gets$ \textit{Selection}($\textit{Pop}^g,P_\textit{PGM}$) \COMMENT{binary tournament}
    \STATE {$\textit{PGM}\gets$ \textit{ProbabilisticModelEstimation}($\textit{Pop}_\mathrm{PGM}^g$)}

    \LCOMMENT{EDA: sampling}
    \STATE $\textit{Pop}_\mathrm{smp}\gets $ \textit{Sampling}($\textit{PGM}, P_\textit{smp}$)
    \FOR{each solution $\mathbf{x} \in \textit{Pop}_\mathrm{smp}$}
      \STATE $\textit{fitness}(\mathbf{x})\gets $\textit{EvaluateFitness}($\mathbf{x},I$)
    \ENDFOR
    \STATE $\textit{Pop}_\mathrm{smp} \gets$ \textit{ParetoDominance}($\textit{Pop}_\mathrm{smp}$);

    \LCOMMENT{EDA: survival}
    \STATE$\textit{Pop}^{g+1}\gets $\textit{Selection}($\{\textit{Pop}^{g} \cup \textit{Pop}_\mathrm{smp}\},P$) \COMMENT{truncation selection}

    \STATE $g \gets g+1$
\UNTIL{no success and $\textit{T}_\textit{max}$ is not exceeded}

\STATE {$\textit{Pop}_\mathrm{ND}\gets  \textit{Pop}^{g-1}(\mathbf{x})$};
\end{algorithmic}
\end{algorithm}

In the context of the adressed MNK-landscape problem, the \emph{Initialization} phase loads the problem instance for a given $M$, $N$ and $K$ (both the subfunctions and the neighborhood structure are obtained from a uniform distribution) and randomly generates an initial population $Pop^1$
of $P$ solutions. Each solution $\mathbf{x}$ is a binary string of size $N$.

The \emph{EvaluateFitness} phase, Step $6$ in Algorithm~\ref{alg:mobeda_framework}, calculates the fitness based on the MNK-landscape model objective functions.

In the \emph{ParetoDominance} phase, the individuals are sorted using Non-dominated Sorting~\cite{srinivas:94} and a binary tournament selects $N_{PGM}$ individuals from $Pop^g$ in the \emph{Selection} phase. The procedure randomly selects two solutions and the one positioned in the best front is chosen. If they lie in the same front, it chooses that solution with the greatest crowding distance. Then, $Pop_{PGM}^g$ is obtained encompassing $N_{PGM}$ good individuals.

Afterward, the algorithm starts, at Step $9$, the PGM construction phase in \emph{ProbabilisticModelEstimation},
according to $Pop_\mathrm{PGM}^g$ population.

Aiming to learn the PGM, the network is modeled using the Bayesian estimate (Equation~\ref{eq:est_bayes}) associated with the 
K2 algorithm.

The PGM is used to sample the set of new solutions ($Pop_\mathrm{smp}$) in Step $10$.
New solutions (a total of $N_{smp}$), are generated from the joint distribution encoded by the network using the probabilistic logic sampling.

Solutions from $Pop_\mathrm{smp}$ are then evaluated and sorted according to the \emph{ParetoDominance}. The sampled population ($Pop_\mathrm{smp}$) is joined with $Pop^g$ to create the new population for the next generation.
However, only the $P$ best solutions are selected (truncation selection) in the \emph{Survival}
process to proceed in the evolutionary process as a new population $Pop^{g+1}$.

This process is iteratively performed until a termination criterion is satisfied.
In this paper, such as in~\cite{liefooghe2015feature}, 
we are interested in the runtime, in terms of a number of function evaluations, until a $(1+\epsilon)$-approximation of the Pareto set is identified (success), subject to a maximum budget of function evaluations ($T_{max}$) for each run.

\subsection{Estimation of the Expected Runtime ($ert$)}

Consider $\epsilon$ as a constant value where $\epsilon\ge0$. For $\bf{x},\bf{x^{'}}\in X$,
$\bf{x}$ is $\epsilon$-dominated by $\bf{x^{'}}$ ($\bf{x}\weakdom_{\epsilon}\bf{x^{'}}$) iff $f_m(\bf{x})\le (1+\epsilon)*f_m(\bf{x^{'}})$, $\forall m \in \{1,...,M\}$.  A set $X^{\epsilon} \subseteq X$ is an $(1 + \epsilon)$-approximation
of the Pareto set if for any solution $\bf{x}\in X$, there is one solution
$\bf{x^{'}}\in X^{\epsilon}$ such that $\bf{x}\weakdom_{\epsilon}\bf{x^{'}}$. This is equivalent to finding
an approximation set whose multiplicative epsilon quality indicator
value with respect to the (exact) Pareto set is lower than $(1+\epsilon)$~\cite{liefooghe2015feature,daolio2015global}.

In order to measure algorithm performance (search cost) In this work we use the expected number of function evaluations necessary to achieve a $(1+\epsilon)$-approximation. We apply the same approach presented in~\cite{liefooghe2015feature,daolio2015global}: we record the number of function evaluations until a $(1+\epsilon)$-approximation is found which characterizes success. Otherwise, the search cost is set to $T_{max}$,

We consider that the algorithm has a
probability of success $p_s \in (0; 1]$ and define
$T_f$ as the random variable measuring the "simulated runtime" (number of function evaluations) for unsuccessful runs (failures). Precisely, after $(t -1)$ failures, each one
requiring $T_f$ evaluations, and the final successful run of $T_s$ evaluations,
the total runtime is $T=\sum_{i=1}^{t-1}T_f+T_s$, where $t$ is the random variable measuring
the number of runs. The random variable $t$ follows a geometric
distribution with parameter $p_s$. 

By taking the expectation and by considering independent runs for each
instance, stopping at the first success, we have:

\begin{equation}
\mathbf{E}[T]=(\mathbf{E}[t]-1)\mathbf{E}[T_f]+\mathbf{E}[T_s]
\end{equation}

In our case, the estimated success rate ($\hat{p}_s$) is computed by the ratio of successful runs over the total number of executions, considering the property that the expectation of a geometric distribution for $t$ with parameter $p_s$ is equal to $1/p_s$. The expected runtime for unsuccessful runs $\mathbf{E}[T_f]$ is set as a constant limit ($T_{max}$)
on the number of function evaluation calls, and the expected runtime for successful runs $\mathbf{E}[T_s]$ is estimated as the average number of function evaluations performed by successful runs. Therefore $ert$ can be expressed as an estimation of the expected runtime $\mathbf{E}[T]$~\cite{liefooghe2015feature,daolio2015global}:

\begin{equation}
ert=\frac{1-\hat{p}_s}{\hat{p}_s}T_{max}+\frac{1}{t_s}\sum_{i=1}^{t_s}T_i
\end{equation}
where $t_s$ is the number of successful runs, $T_i$ is the number of
evaluations for successful run $i$.

\section{Experiments and Results} \label{sec:experiments}

In this section, we are interested in the ability of the \algName, presented in Section \ref{sec:edas}, to find a Pareto set approximation for multi and many-objective combinatorial optimization problems in comparison with NSGA-III~\cite{deb:14}, a state-of-the-art algorithm applied to solve MOPs. In particular, we investigate the (estimated) runtime of \algName~and NSGA-III necessart to identify a $(1+\epsilon)-$approximation of the Pareto set over enumerable MNK-landscapes instances. 

We consider a population size of $P=100$ for both algorithms. For \algName~the number of solutions selected to support the probabilistic model estimation is $P_{PGM}=P/2$, and the number of solutions sampled from the probabilistic model is $P_{smp}=10*P$.

NSGA-III was adapted from PlatEmo platform~\cite{tian2017platemo} considering Uniform Crossover Probability of $0.8$ and Bit Flip Mutation Probability of $1/500$, as well the same method for the number of reference points used by~\cite{deb:14}. 

We consider MNK-landscapes with an epistatic degree $K \in \{2, 4, 6, 8, 10\}$, an objective space dimension $M \in \{2, 3, 5, 8\}$.  The problem size is set to $N = 18$ in order to enumerate the solution space exhaustively - we used the largest value of $N$ that can still be analyzed with reasonable computational resources. A set of $30$ different landscapes are independently generated at random for each parameter combination $M$ and $K$. The time limit is set to $T_{max} =
2^N*10^{-1} < 26 215$ function evaluations~\cite{liefooghe2015feature} without identifying a $(1 + \epsilon)$-approximation. Each algorithm is executed $100$ times per instance, with $\epsilon=0.1$.
The number of neighbors per variable is the same for all functions $f_m$, i.e. $K_m=K$ for all $m \in \{1, \dots, M\}$, as proposed in \cite{aguirre2007working,santana2015evolving}. 

For each landscape, we enumerate the search space classifying solutions into non-dominated fronts. The first front
is the Pareto front and corresponds to the Pareto optimal set that contains the best non-dominated solutions.

\subsection{Fitness Lansdcape Analysis}

In this paper we consider some features extracted from the problem instance (low-level features), or computed from the enumerated Pareto set and solution space (high-level features)~\cite{liefooghe2015feature}, as presented in Table~\ref{tab:features}. 
For more details and a comprehensive explanation of these features, the reader is referred to~\cite{liefooghe2013makes}.
We addressed these features in order to examine their impact on the algorithms performance.
Note that for the hypervolume computation, the reference point is set to the origin.

\begin{table}[ht]
  \scriptsize
  \centering
  \caption{Low-level and high-level features used in the study}\label{tab:features}
  \vspace{-.2cm}
  \begin{tabular}{|l l l|}
    \multicolumn{3}{c}{low-level features} \\
    \hline
    k	&Number of variable interactions &	\\
    m &Number of objective functions & \\
    \hline
    \multicolumn{3}{c}{high-level features} \\
    \hline
    npo	&Number of Pareto optimal solutions &\cite{aguirre2007working}	\\
    hv &Hypervolume value of a the Pareto set & \cite{aguirre2007working}\\
    avgd &Average distance between Pareto optimal solutions & \cite{liefooghe2013makes} \\
    maxd &Maximum distance between Pareto optimal solutions & \cite{liefooghe2013makes}\\
    nconnec &Number of connected components in the Pareto set & \cite{liefooghe2013local} \\
    lconnec &Proportion of the largest connected component of the Pareto set & \cite{liefooghe2013local}\\
    kconnec &Minimal Hamming distance to connect the Pareto set & \cite{liefooghe2013local} \\
    \hline
  \end{tabular}
\end{table}

As an attempt to understand the impact of problem features on both \algName~and NSGA-III performances, we conduct a linear regression analysis on the correlation between the problem features presented in Table~\ref{tab:features} and the estimated runtime ($ert$). 

The cost model for the linear regression adopted in this paper is formalized in equation \ref{eq:lin-reg}.
\begin{equation}\label{eq:lin-reg}
ert=\beta_0+\beta_1.v_1+\beta_2.v_2+...+\beta_p,v_p+e
\end{equation}

The response variable ($ert$) is explained by $p=9$ variables corresponding to the selected problem features shown in Table \ref{tab:features}. The response is log-transformed in order to better approach linearity.

\begin{figure*}[htbp]
  \centering
  \subfloat{
    \includegraphics[scale=0.36]{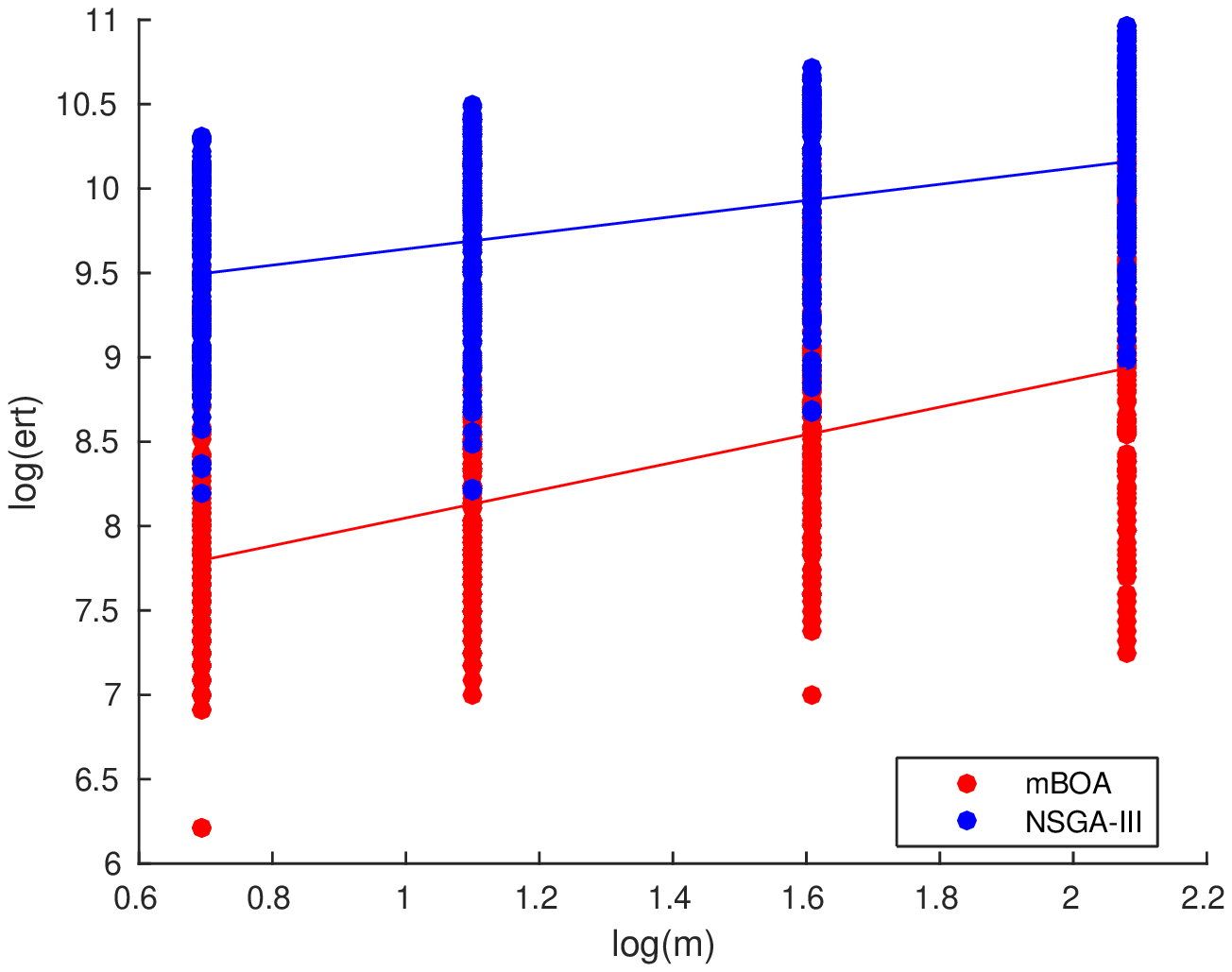}
    \label{fig:mXert}
  }
  \quad 
  \subfloat{
    \centering
    \includegraphics[scale=0.36]{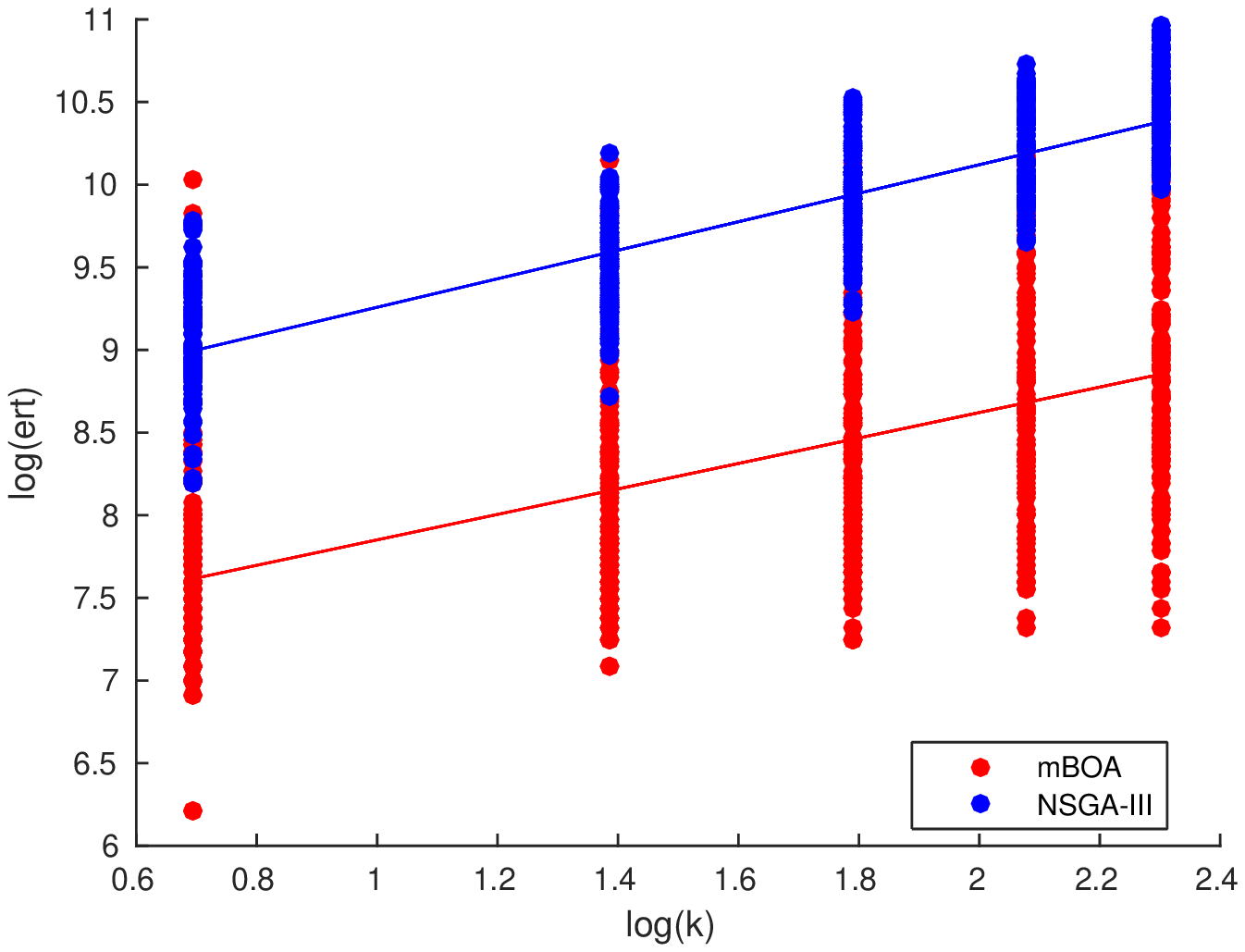}
    \label{fig:kXert}
  }
  \quad 
  \subfloat{
    \centering
    \includegraphics[scale=0.36]{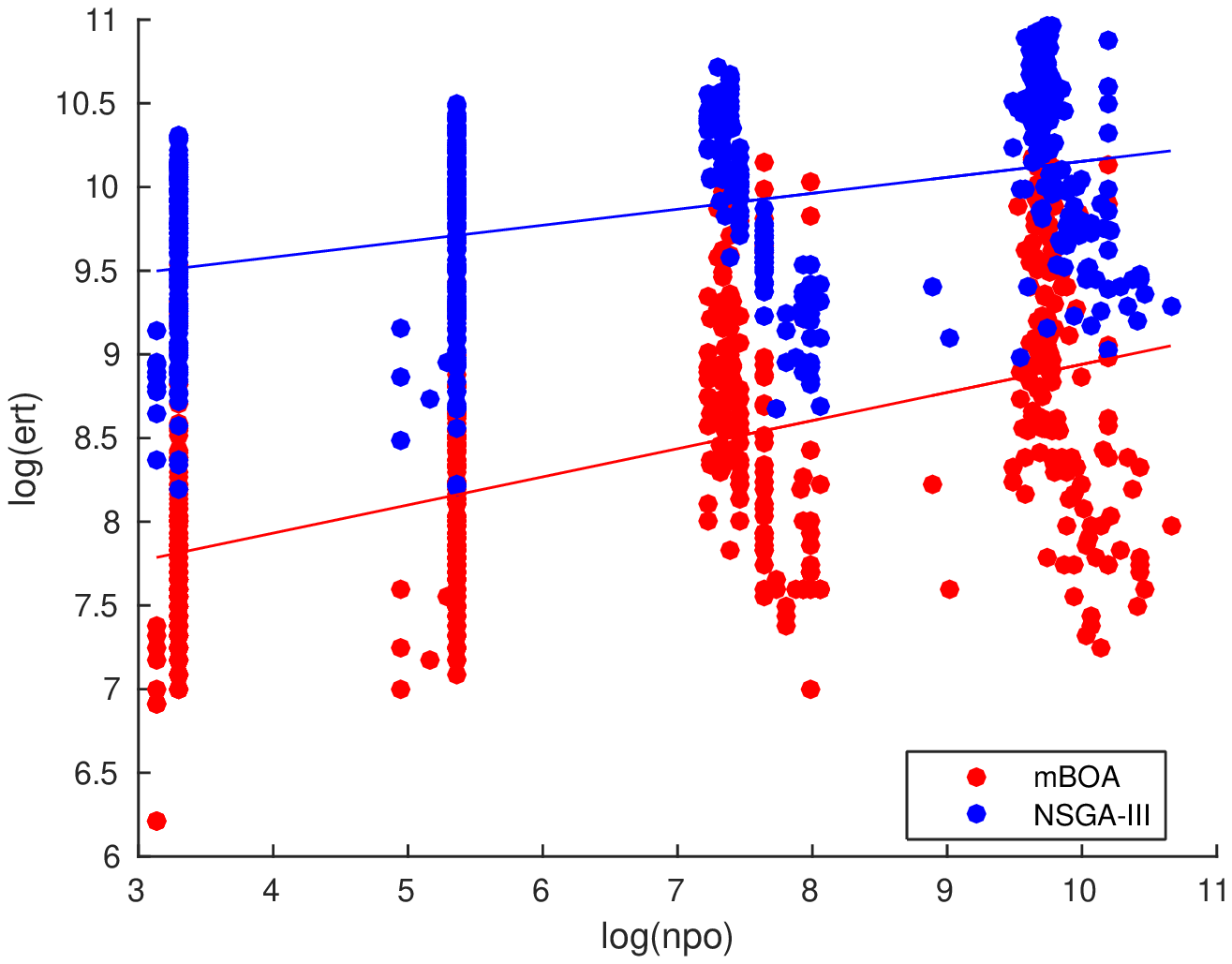}
    \label{fig:npoXert}
  }
  \quad 
  \subfloat{
    \centering
   \includegraphics[scale=0.36]{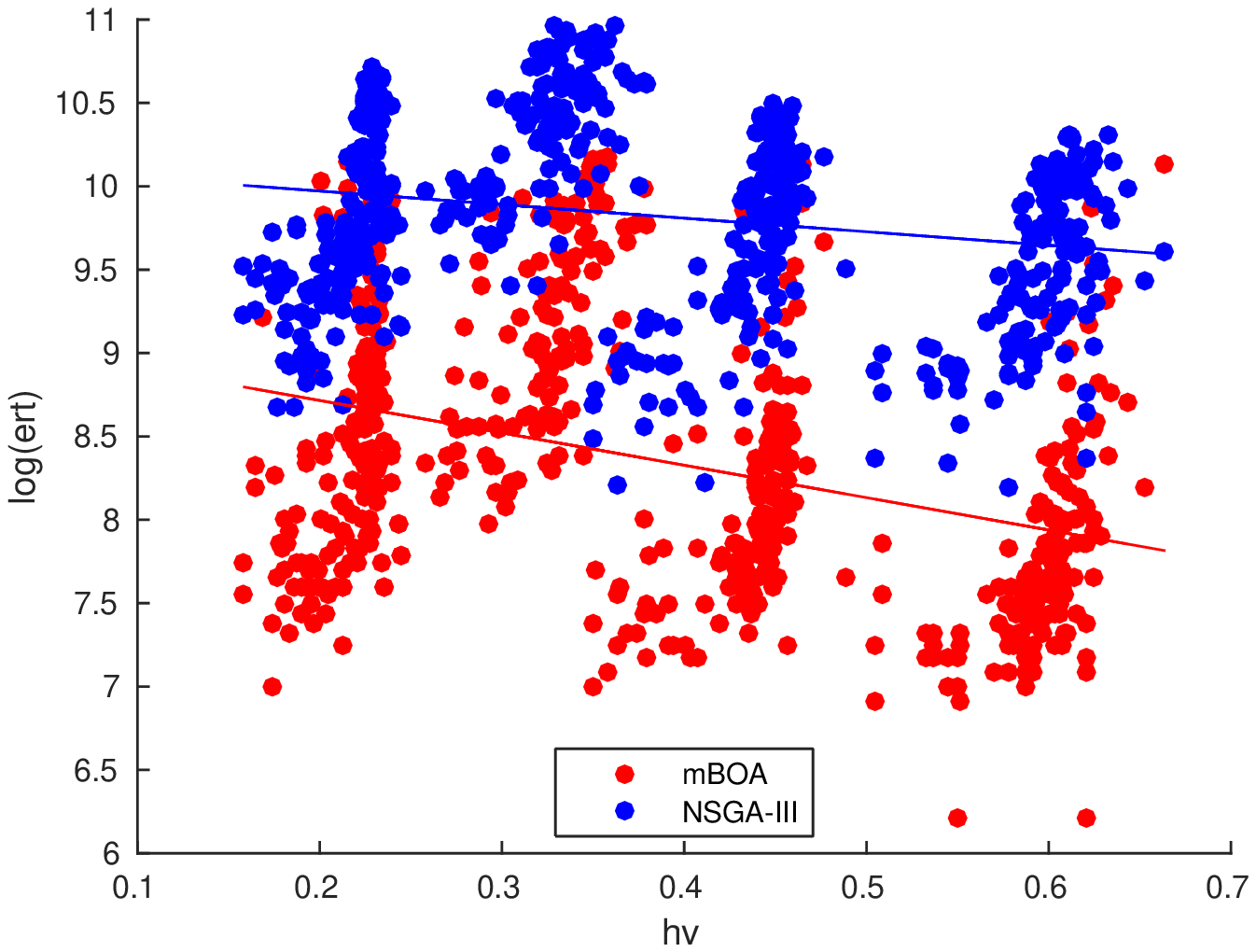}
    \label{fig:hvXert}
  }
  \quad 
  \subfloat{
    \centering
    \includegraphics[scale=0.36]{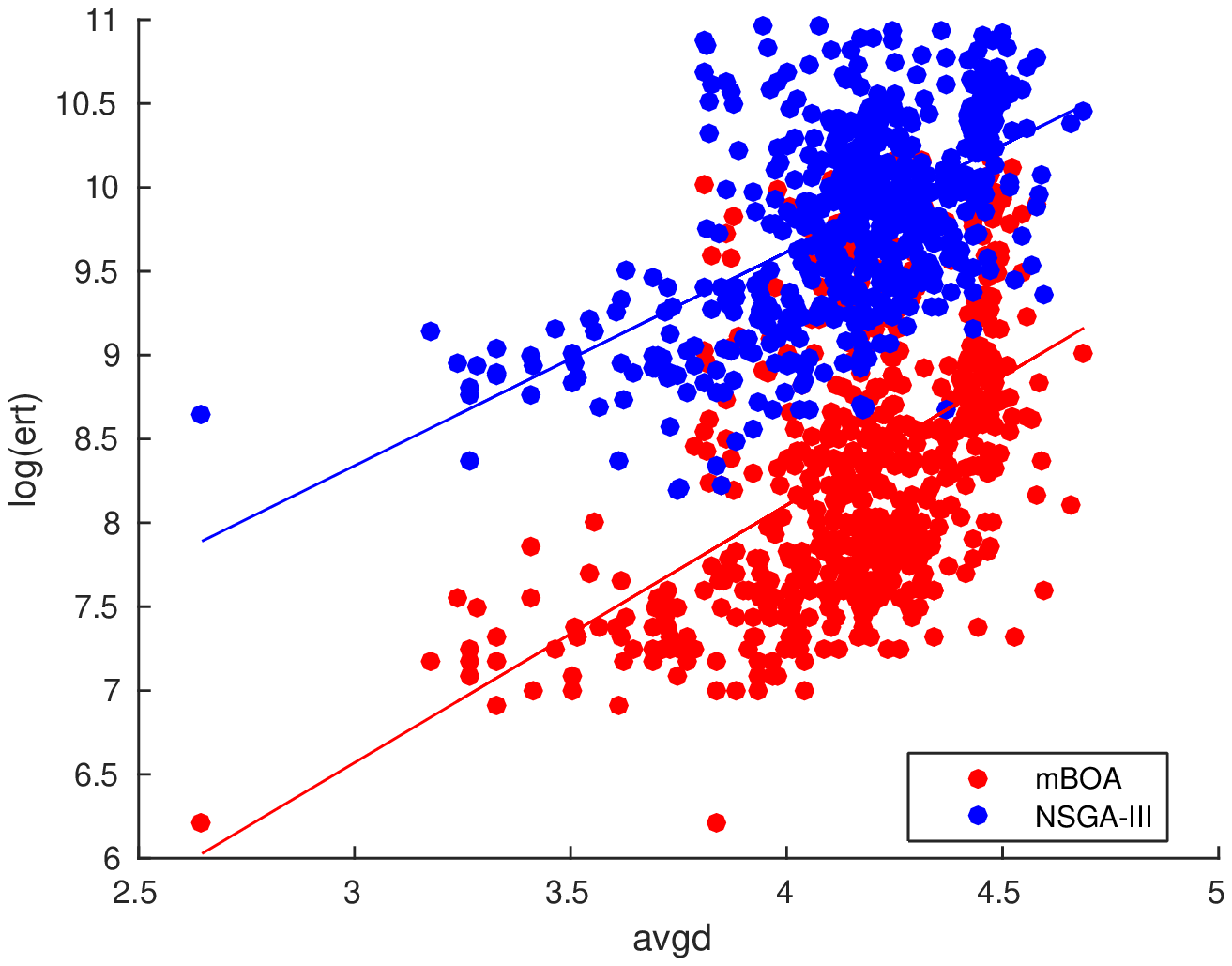}
    \label{fig:avgdXert}
  }
  \quad 
  \subfloat{
    \centering
    \includegraphics[scale=0.36]{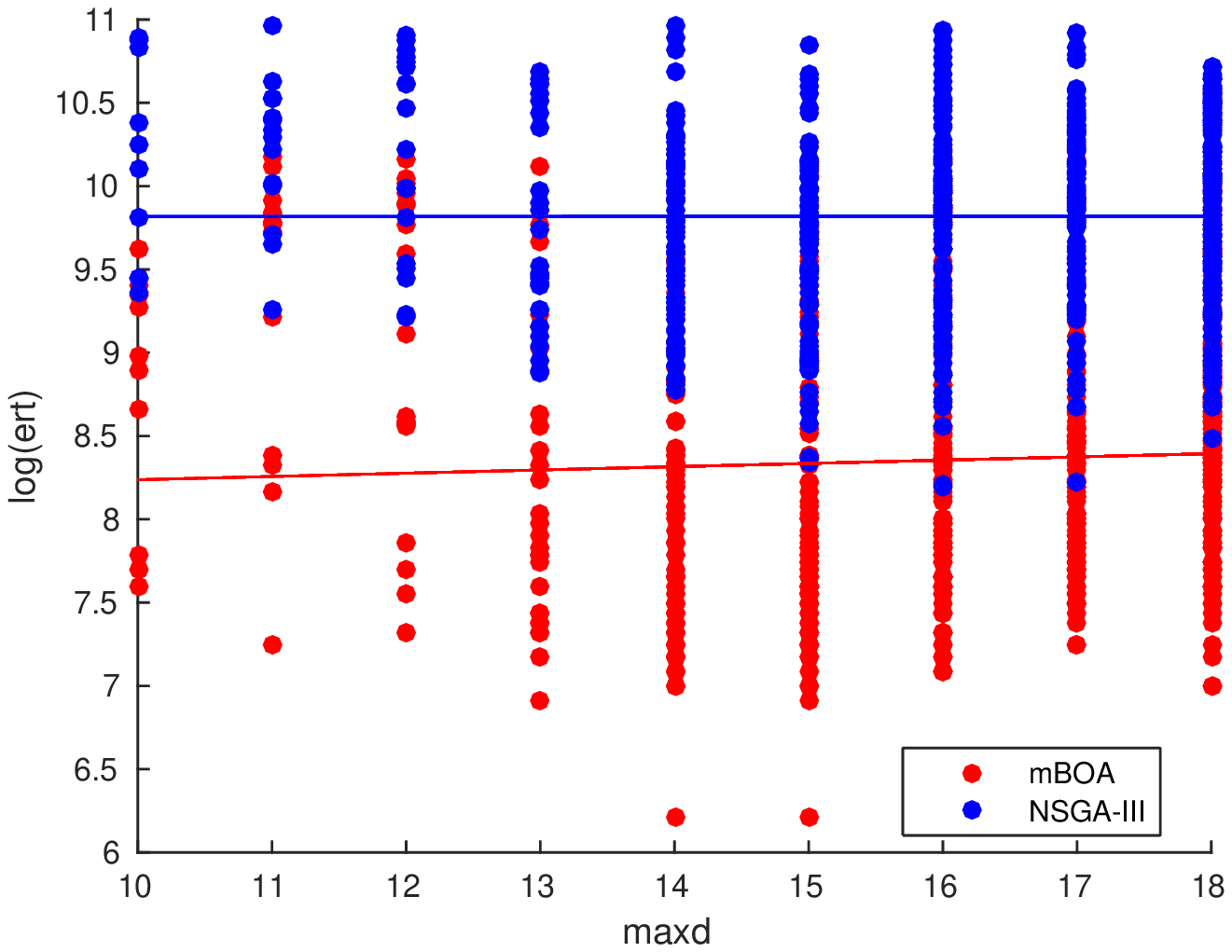}
    \label{fig:maxdXert}
  }
  \quad 
  \subfloat{
    \centering
    \includegraphics[scale=0.36]{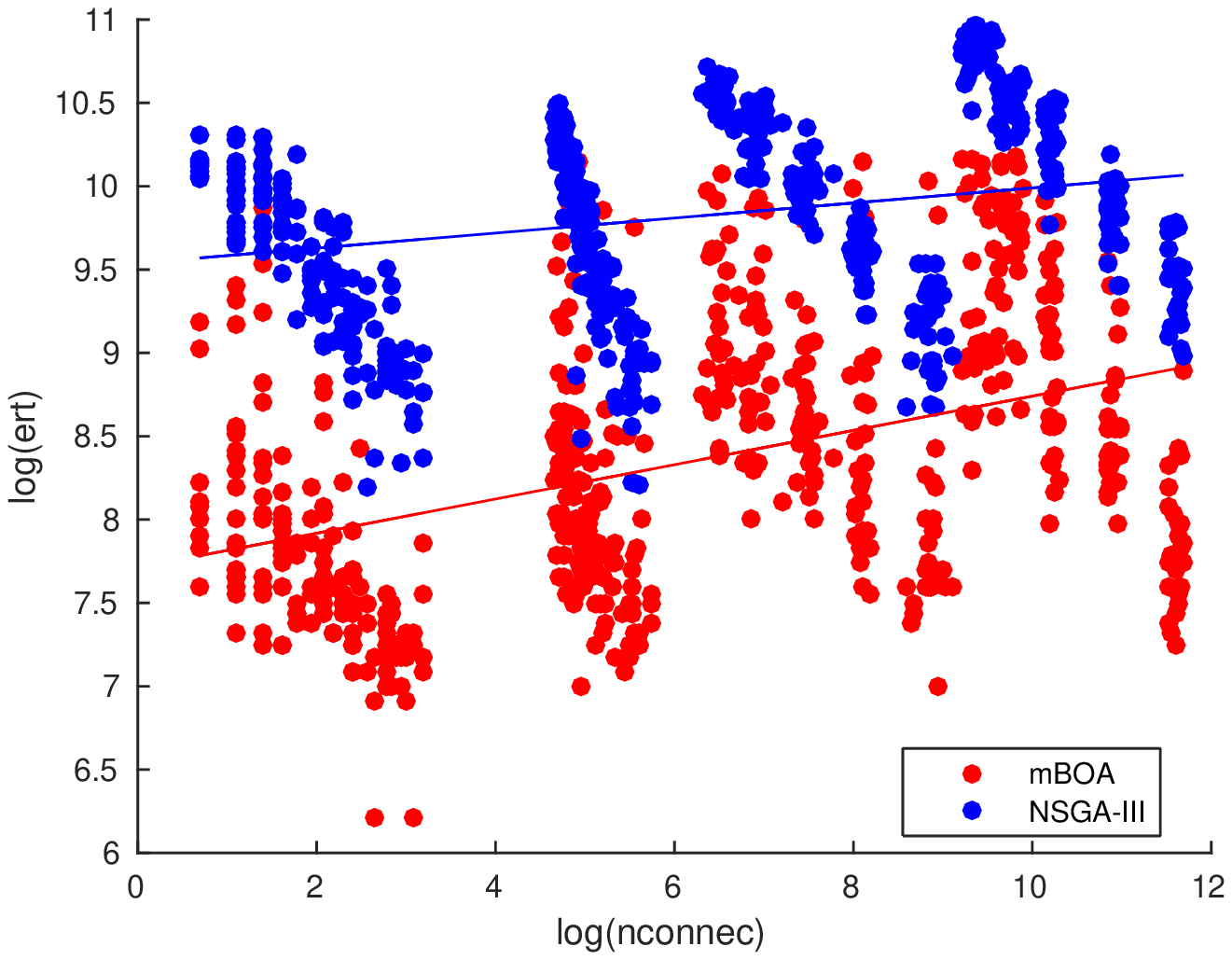}
    \label{fig:nconnecXert}
  }
  \quad 
  \subfloat{
    \centering
    \includegraphics[scale=0.36]{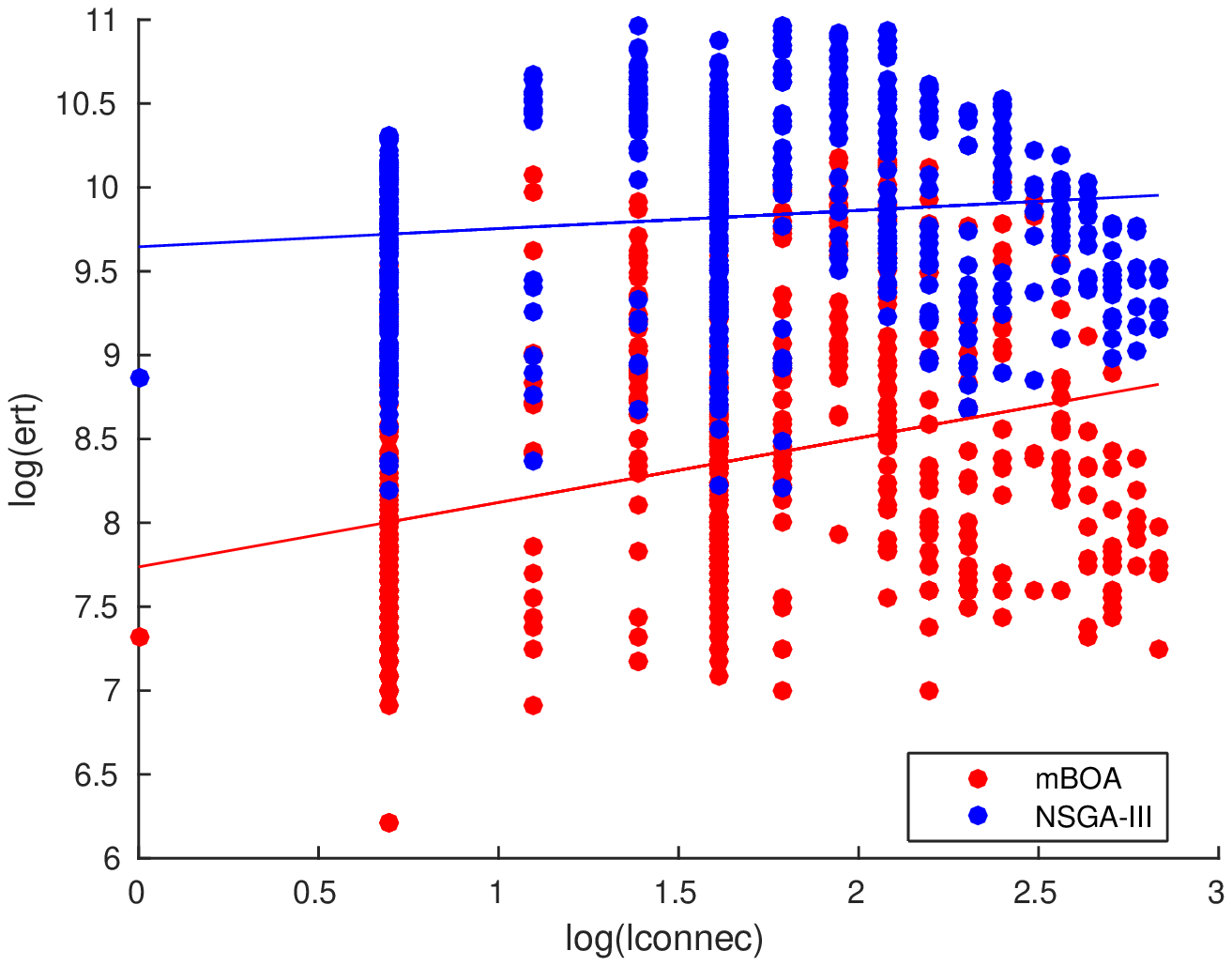}
    \label{fig:lconnecXert}
  }
  \quad 
  \subfloat{
    \centering
    \includegraphics[scale=0.36]{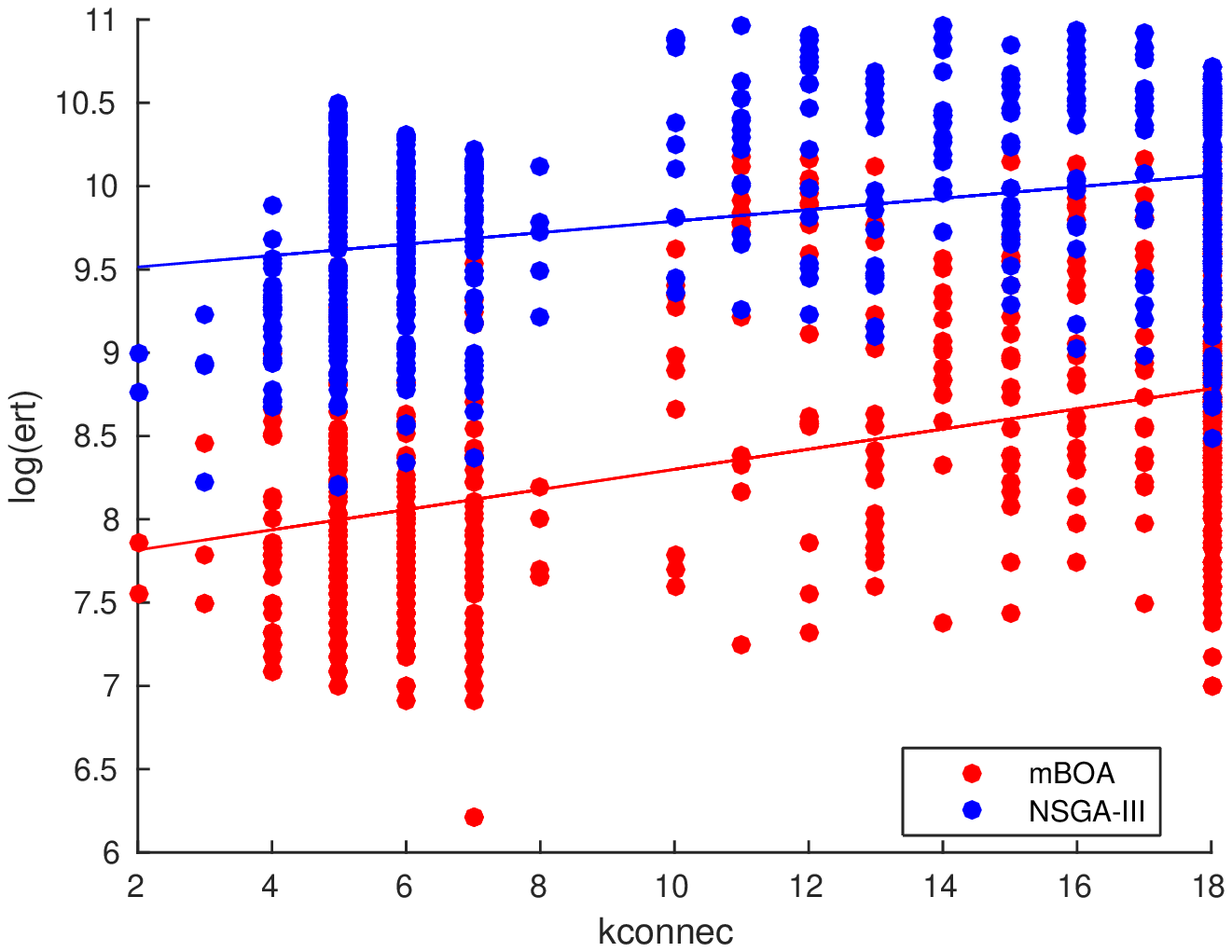}
    \label{fig:kconnecXert}
  }
  \vspace{-.2cm}
  \caption{Scatter plots and regression lines for each feature vs. log(ert).}
  \label{fig:featuresXert}
  \vspace{-.4cm}
\end{figure*}

Figure \ref{fig:featuresXert} represents the scatter plots of multiple problem features and the log-transformed $ert$ for MOEDA and NSGA-III, in addition to the regression lines.

To measure the accuracy of the linear regression model, the following statistics are analyzed in the experiments~\cite{liefooghe2015feature}:

\begin{itemize}
  \item The absolute correlation coefficient $r$ measures the linear association between the predicted and the actually observed values (the Pearson correlation coefficient is used here). Its absolute value ranges from $0$ to $1$. Values closer to $1.0$ mean better fittings.
  \item The mean absolute error $MAE$ corresponds to the average value of the absolute difference between the values predicted by the regression model and the values actually observed (i.e. the residuals), therefore, the lower the $MAE$, the better the regression model.
  \item The root mean-square error $RMSE$ measures the square root of the average squared difference between the values predicted by the regression model and the values actually observed. The lower the $RMSE$, the better the regression model.
\end{itemize}

The corresponding regression model statistics are reported in Table \ref{tab:feature-stats}. Aiming at approaching linearity, we also use a log-log scale for the features $m$, $k$, $npo$, $nconnect$, and $lconnect$. Note that the amount of bias in the set of results is negligible according to the equivalent statistics obtained using 10-fold cross validation. 

\begin{table}[ht]
  \centering
  \caption{Statistics of the simple linear regression models.}
  \label{tab:feature-stats}
  \vspace{-.2cm}

  \begin{tabular}{l c c c | c c c }
    \hline
    \multicolumn{4}{c}{Linear Regression - mBOA} & \multicolumn{3}{c}{10-fold cross validation}\\
    \hline
                  & r        & MAE     & RMSE     &  r        & MAE    & RMSE   \\
    none          &$ 0.00   $&$ 0.68  $&$ 0.82  $ & $  0.00  $&$ 0.68 $&$ 0.82 $\\ \hline
    log(lconnec)  &$ 0.30   $&$ 0.65  $&$ 0.78  $ & $  0.30  $&$ 0.65 $&$ 0.79 $\\
    hv            &$  0.35  $&$ 0.62  $&$ 0.77  $ & $  0.30  $&$ 0.64 $&$ 0.79 $\\
    log(nconnec)  &$  0.40  $&$ 0.61  $&$ 0.75  $ & $  0.40  $&$ 0.62 $&$ 0.76 $\\
    kconnec       &$ 0.41   $&$ 0.60  $&$ 0.75  $ & $  0.40  $&$ 0.60 $&$ 0.75 $\\
    maxd          &$  0.48  $&$ 0.68  $&$ 0.82  $ & $  0.46  $&$ 0.68 $&$ 0.83 $\\
    avgd          &$  0.48  $&$ 0.58  $&$ 0.72  $ & $  0.48  $&$ 0.59 $&$ 0.73 $\\
    log(npo)      &$  0.50  $&$ 0.57  $&$ 0.71  $ & $  0.50  $&$ 0.57 $&$ 0.72 $\\
    log(m)        &$  0.52  $&$ 0.56  $&$ 0.70  $ & $  0.53  $&$ 0.56 $&$ 0.71 $\\
    log(k)        &$  0.53  $&$ 0.55  $&$ 0.69  $ & $  0.54  $&$ 0.56 $&$ 0.70 $\\ \hline

    \multicolumn{4}{c}{Linear Regression - NSGA-III} & \multicolumn{3}{c}{10-fold cross validation}\\ \hline
                  & r        & MAE     & RMSE     &  r        & MAE    & RMSE   \\
    none          &$ 0.00   $&$ 0.48  $&$ 0.58  $ & $  0.00  $&$ 0.49 $&$ 0.58 $\\ \hline
    log(lconnec)  &$ 0.12   $&$ 0.48  $&$ 0.57  $ & $  0.12  $&$ 0.45 $&$ 0.57 $\\
    hv            &$  0.21  $&$ 0.48  $&$ 0.57  $ & $  0.22  $&$ 0.48 $&$ 0.58 $\\
    log(nconnec)  &$  0.25  $&$ 0.48  $&$ 0.56  $ & $  0.24  $&$ 0.48 $&$ 0.56 $\\
    kconnec       &$ 0.33   $&$ 0.46  $&$ 0.55  $ & $  0.32  $&$ 0.47 $&$ 0.56 $\\
    maxd          &$  0.40  $&$ 0.48  $&$ 0.58  $ & $  0.40  $&$ 0.48 $&$ 0.59 $\\
    log(npo)      &$  0.40  $&$ 0.46  $&$ 0.53  $ & $  0.41  $&$ 0.47 $&$ 0.53 $\\
    log(m)        &$  0.43  $&$ 0.45  $&$ 0.52  $ & $  0.44  $&$ 0.45 $&$ 0.53 $\\
    avgd          &$  0.57  $&$ 0.38  $&$ 0.48  $ & $  0.56  $&$ 0.39 $&$ 0.48 $\\
    log(k)        &$  0.85  $&$ 0.25  $&$ 0.31  $ & $  0.86  $&$ 0.25 $&$ 0.31 $\\ \hline
  \end{tabular}
   \vspace{-.2cm}
\end{table}

Figure \ref{fig:featuresXert} shows that \algName~has a significantly lower estimated runtime compared to NSGA-III. 
The average $ert$ values for \algName~and NSGA-III are respectively $5922$  and $21498$ with respective standard deviations of $5682$ and $11696$. 
The statistics presented in Table \ref{tab:feature-stats} show that the feature that has the most significant impact is the ruggedness. This can be clearly seen in NSGA-III for which we obtained a high correlation of $0.85$ between $log(k)$ and $log(ert)$.

On the other hand, most features seem to have a moderate or low impact on the \algName performance. The correlation of features and $ert$ is bounded between $0.30$ and $0.53$, with the low-level features $k$ and $m$ being those with the highest impact. 

Given the above-mentioned remarks, it is hard to use a linear regression model to predict an algorithm performance based on the individual problem features. However, it is possible to obtain reasonable precision of runtime prediction using the ruggedness or the number of objectives. In fact, the $MAE$ values of \algName~for $k$ and $m$ are respectively $0.55$ and $0.56$, which are slightly better than the precisions of the other models. The fact that the $RMSE$ values ($0.70$ and $0.69$ respectively) are larger than the corresponding $MAE$ values suggests that there is error variation. However, the gap between $RMSE$ and $MAE$ is not large enough, indicating that large errors are unlikely to happen.

\begin{table}[ht]
  \centering
  \caption{Statistics for the multiple linear regression models using feature elimination}
  \label{tab:multi-linear-regression}
  \vspace{-.2cm}

  \begin{tabular}{l c c c | c c c} \hline
    \multicolumn{4}{c}{Multiple Linear Regression - mBOA} & \multicolumn{3}{c}{10-fold cross validation}\\ \hline
                                  & r      & MAE    & RMSE   & r      & MAE    & RMSE   \\
    all                           &$ 0.86 $&$ 0.20 $&$ 0.22 $&$ 0.83 $&$ 0.20 $&$ 0.22 $\\ \hline
    $\smallsetminus$ maxd         &$ 0.87 $&$ 0.20 $&$ 0.22 $&$ 0.84 $&$ 0.21 $&$ 0.22 $\\
    $\smallsetminus$ npo          &$ 0.87 $&$ 0.21 $&$ 0.22 $&$ 0.84 $&$ 0.21 $&$ 0.23 $\\
    $\smallsetminus$ avgd         &$ 0.87 $&$ 0.21 $&$ 0.22 $&$ 0.84 $&$ 0.22 $&$ 0.23 $\\
    $\smallsetminus$ kconnect     &$ 0.88 $&$ 0.22 $&$ 0.23 $&$ 0.85 $&$ 0.22 $&$ 0.24 $\\
    $\smallsetminus$ log(lconnec) &$ 0.88 $&$ 0.22 $&$ 0.23 $&$ 0.86 $&$ 0.23 $&$ 0.24 $\\
    $\smallsetminus$ hv           &$ 0.89 $&$ 0.22 $&$ 0.23 $&$ 0.86 $&$ 0.23 $&$ 0.24 $\\
    $\smallsetminus$ log(nconnec) &$ 0.89 $&$ 0.23 $&$ 0.24 $&$ 0.87 $&$ 0.23 $&$ 0.25 $\\
    $\smallsetminus$ log(m)       &$ 0.39 $&$ 0.51 $&$ 0.69 $&$ 0.38 $&$ 0.53 $&$ 0.72 $\\
    $\smallsetminus$ log(k)       &$ 0.00 $&$ 0.55 $&$ 0.70 $&$ 0.00 $&$ 0.57 $&$ 0.72 $\\ \hline

    \multicolumn{4}{c}{Multiple Linear Regression - NSGA-III} & \multicolumn{3}{c}{10-fold cross validation}\\ \hline
                                  & r      & MAE    & RMSE   & r      & MAE    & RMSE   \\
    all                           &$ 0.95 $&$ 0.14 $&$ 0.17 $&$ 0.92 $&$ 0.14 $&$ 0.17 $\\ \hline
    $\smallsetminus$ maxd         &$ 0.95 $&$ 0.14 $&$ 0.17 $&$ 0.92 $&$ 0.14 $&$ 0.17 $\\
    $\smallsetminus$ log(nconnec) &$ 0.95 $&$ 0.14 $&$ 0.18 $&$ 0.92 $&$ 0.14 $&$ 0.18 $\\
    $\smallsetminus$ log(lconnec) &$ 0.95 $&$ 0.14 $&$ 0.18 $&$ 0.92 $&$ 0.14 $&$ 0.18 $\\
    $\smallsetminus$ kconnect     &$ 0.95 $&$ 0.14 $&$ 0.18 $&$ 0.92 $&$ 0.14 $&$ 0.18 $\\
    $\smallsetminus$ avgd         &$ 0.95 $&$ 0.14 $&$ 0.18 $&$ 0.92 $&$ 0.14 $&$ 0.18 $\\
    $\smallsetminus$ hv           &$ 0.95 $&$ 0.14 $&$ 0.18 $&$ 0.92 $&$ 0.14 $&$ 0.18 $\\
    $\smallsetminus$ log(npo)     &$ 0.95 $&$ 0.14 $&$ 0.18 $&$ 0.92 $&$ 0.14 $&$ 0.18 $\\
    $\smallsetminus$ log(m)       &$ 0.43 $&$ 0.45 $&$ 0.52 $&$ 0.42 $&$ 0.46 $&$ 0.53 $\\
    $\smallsetminus$ log(k)       &$ 0.00 $&$ 0.48 $&$ 0.58 $&$ 0.00 $&$ 0.49 $&$ 0.59 $\\ \hline
  \end{tabular}
\end{table}

\begin{figure}[h]
\vspace{-.2cm}
\centering
\quad 
\subfloat[]{
  \centering
  \includegraphics[scale=0.25]{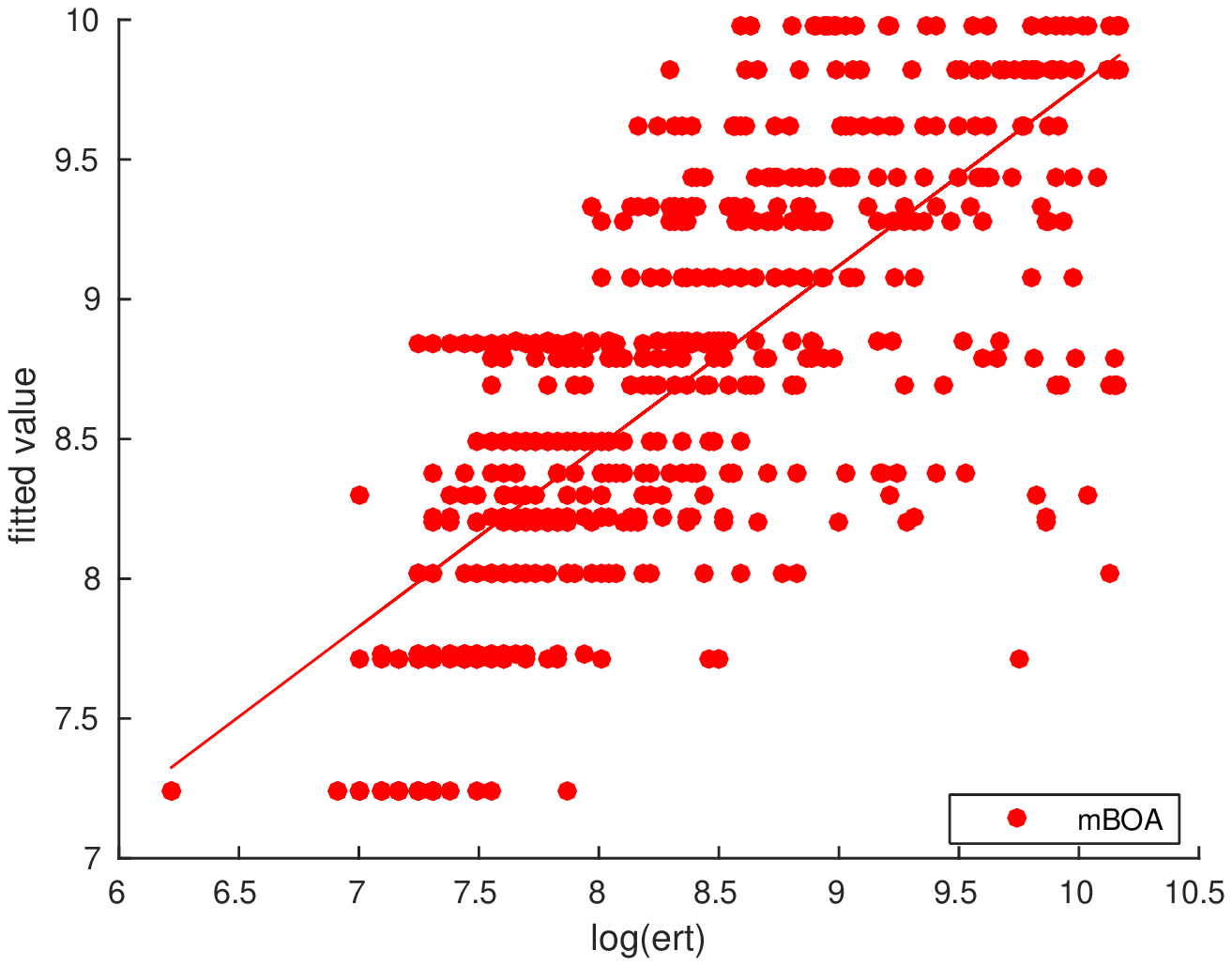}
  \label{fig:kXert}
}
\quad 
\subfloat[]{
  \centering
  \includegraphics[scale=0.25]{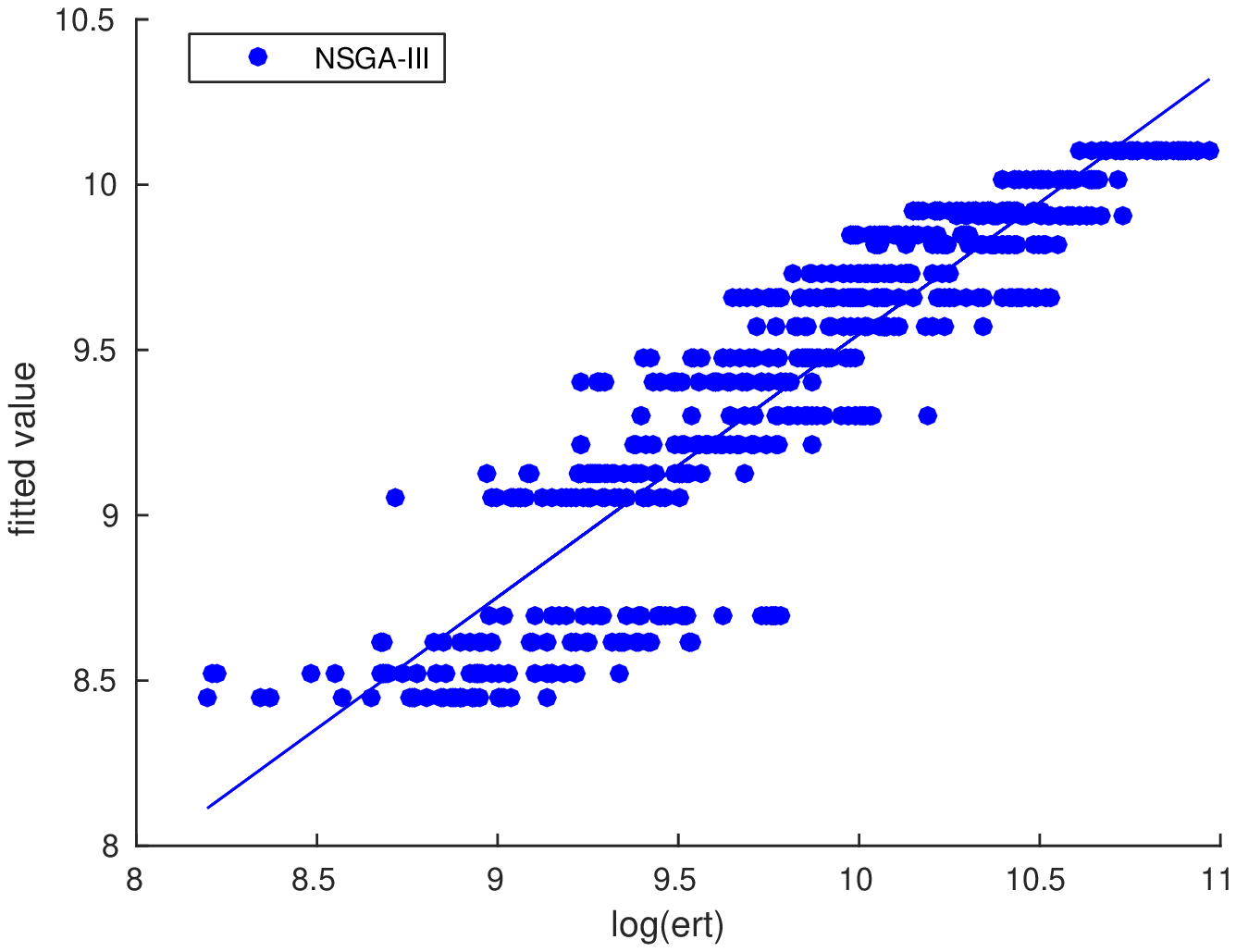}
  \label{fig:hvXert}
}
\vspace{-.2cm}
\caption{log(ert) vs. fitted values for the model with joint features $k$ and $m$ for a) mBOA  and b) NSGA-III.}
\label{fig:subset}
\vspace{-.2cm}
\end{figure}

As the linear model for mBOA does not provide a precise runtime prediction given the individual features, we decided to take our work a step forward by studying the impact of combined problem features using a multiple linear model. To do so, we start by considering the impact of all the problem features on the algorithms $ert$. Since some features may have a negligible or noisy impact, they should be removed from the model. Therefore, we proceed by eliminating the least impact feature at each step, until no feature is left. This process is summarized in Table \ref{tab:multi-linear-regression}, and the corresponding scatter plots are illustrated in Figure \ref{fig:subset}.

The models' statistics presented in Table \ref{tab:multi-linear-regression} show that using combined features provides a significantly more precise model for both algorithms --- specially in mBOA's case. In fact, the multiple linear model
provides a correlation coefficient of $0.89$ instead of the ruggedness-based model, which has a correlation coefficient of $0.53$ for mBOA. We can clearly see that the most influencing features are $k$ and $m$. Indeed, by using only these two feature the runtime of mBOA can be estimated with a relatively small error ($MAE=0.23$ and $RMSE=0.24$).

Interestingly, when removing some features using backward elimination, the correlation coefficient increases. Indeed, the correlation coefficient starts with $0.86$ when using all the features, and increases to $0.89$ after eliminating $6$ features (the first 6 lines in Table \ref{tab:multi-linear-regression}). Although the impact seems too small, it tells us that these features have a negative impact on the efficiency of the model (acting like noisy features).


\subsection{Probabilistic Analysis of Pareto Front}

%

In the previous section we have analyzed the influence of selected features into the estimated runtime for $(1+\epsilon)$ approximations. However, the $(1+\epsilon)$ approximation purely might not reveal some important aspects of non-dominated solutions of the Pareto front as convergence and distribution. These aspects are investigated in this section by analyzing the probabilistic information of the final PGM (BN structure and parameters) for \algName~when the $(1 + \epsilon)-$approximation is found.
We calculate the pmf $P(\mathbf{y})$, defined in Equation \ref{eq:rb}, for each solution in the
Pareto set, based on the PGM learned at the end of each execution of each landscape.

\begin{figure}[h]
  \vspace{-.2cm}
  \centering
  \subfloat{
    \centering
    \includegraphics[scale=.4]{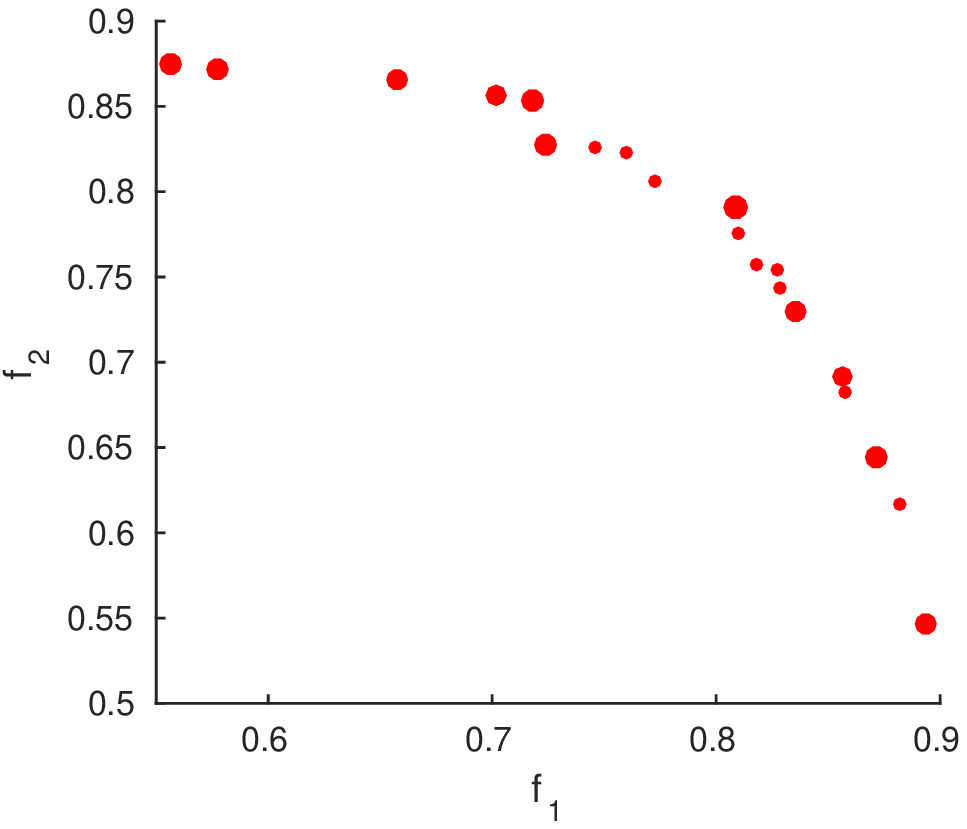}
  }
  \quad 
  \subfloat{
    \centering
    \includegraphics[scale=.4]{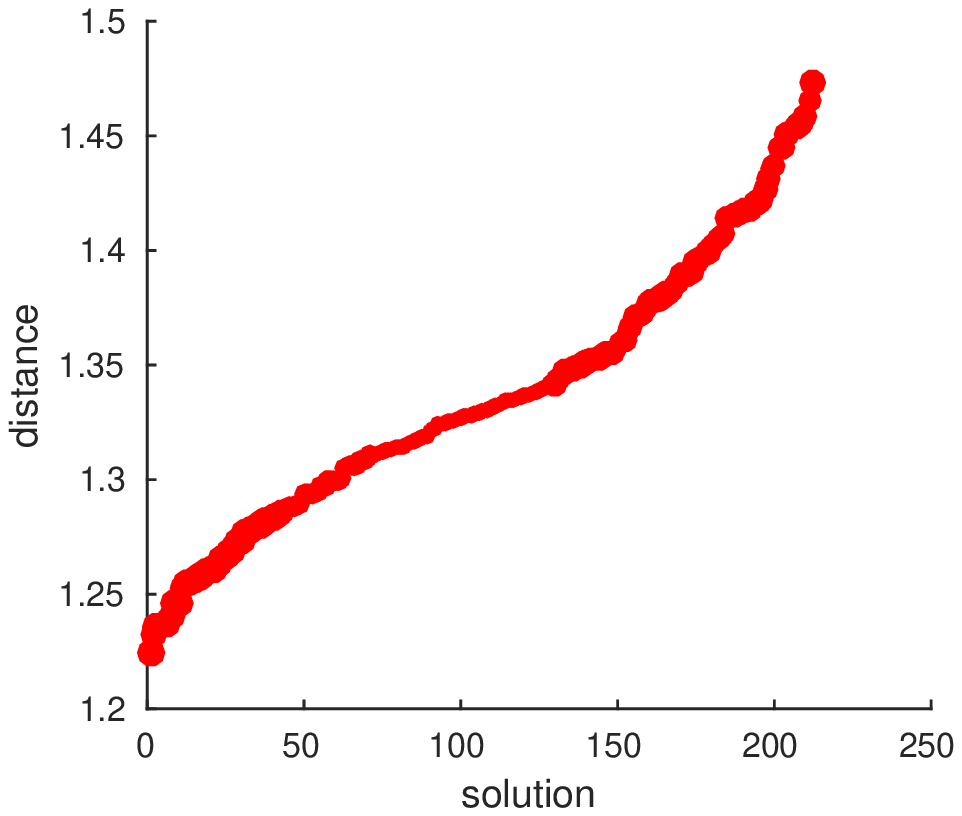}
  }
  \quad 
  \subfloat{
    \includegraphics[scale=.4]{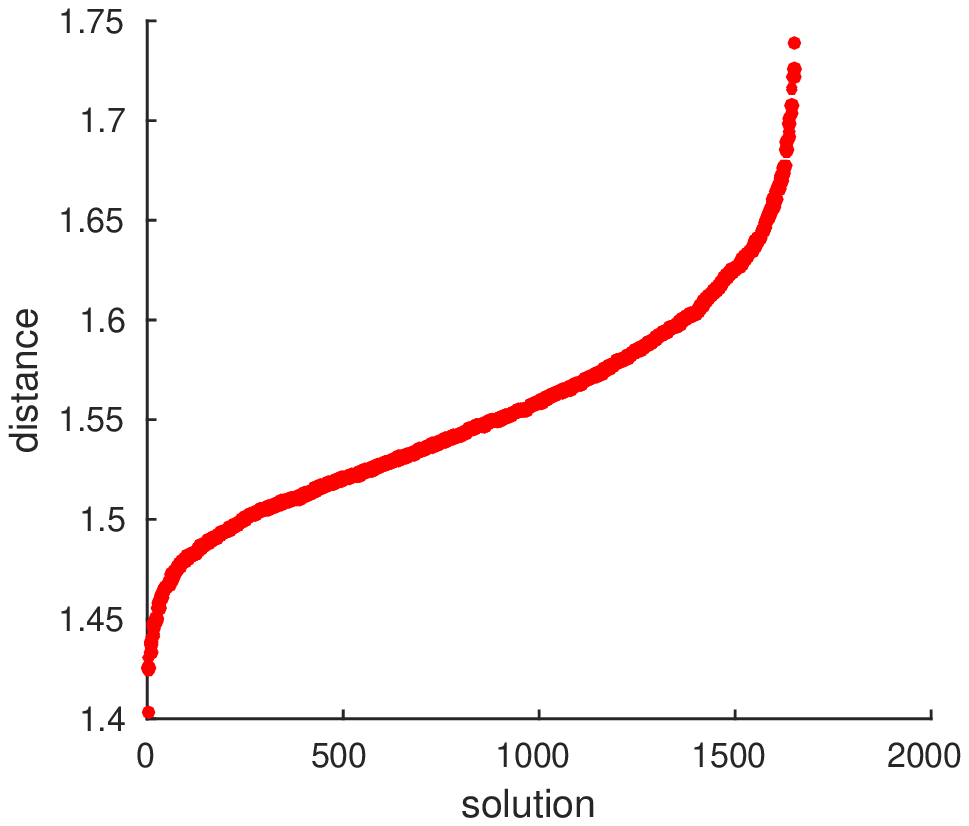}
  }
  \quad 
  \subfloat{
    \centering
    \includegraphics[scale=.4]{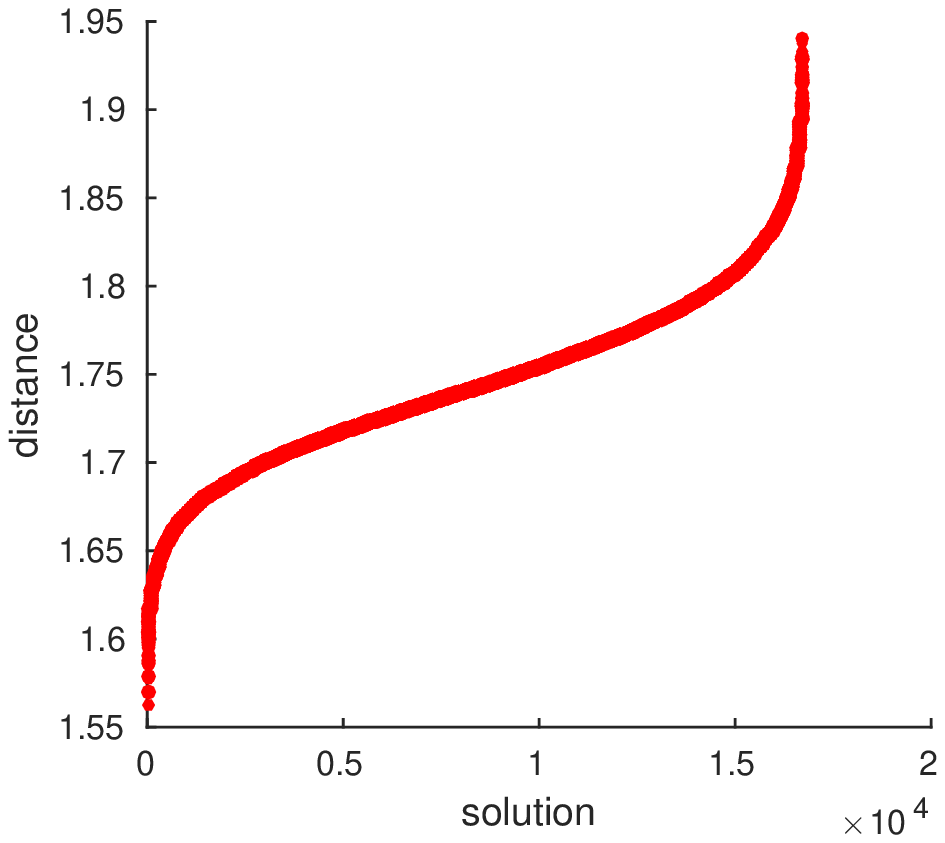}
  }

  \vspace{-.2cm}
  \caption{
    A probabilistic view for the Pareto front for 2 objectives (top-left),
    the Euclidean distance between the ideal point for 3 objectives (top-right), 
    5 objectives (bottom-left), 
    and 8 objectives (bottom-right). 
    $k=6$ in all cases. }
  \label{fig:prob}
  \vspace{-.2cm}
\end{figure}

Afterwards, the mean of the pmf values along all executions is obtained in order to calculate the marginal
distribution $P(Z_1=z_1,...,Z_M=z_M)$. Each non-dominated solution is represented by a circle in its corresponding point in the PF, which is
proportional to the marginal probability $P(Z_1=z_1,...,Z_M=z_M)$~\cite{mar_bracis:17}. Due to the space limitation, we present, in Figure~\ref{fig:prob}, the PF probability view of mBOA for one specific landscape with $K=6$. Since it is not practical to visualize the Pareto front for more than two objectives, we illustrate the ordered Euclidean distance (from the nearest to the farthest) between each point from the Pareto front and the ideal point \footnote{The ideal point is calculated as the maximum value of each objective.} for $3$, $5$, and $8$ objectives.

In the top-left plot in Figure~\ref{fig:prob}, we can see that \algName~shows higher probabilities to the solutions distributed along the entire front.
This is also observed in the top-right plot, however here, some solutions nearby ideal and extreme points present high probabilities.

In the bottom plots for 5 and 8 objectives,  we note that the solutions present similar probabilities, since there are few large points plotted in the Pareto front.

The results show that, for smaller number of objectives,  solutions around Pareto front knee and extreme points are better represented because they have higher probabilities of occurence as depicted Figure~\ref{fig:prob}. This does not happen for 5 and 8 objectives. These observations can be useful to understand the convergence and diversity performance of a given approach. In fact, the experiments we performed show that examining the PGM structures according to the marginal distribution of the corresponding objectives values from the Pareto front can be very useful to analyze the performance of mBOA and in the future we can use this information to guide the search process through specific regions of the Pareto front using the current state of the Bayesian Network.

\section{Conclusion} \label{sec:conclusion}

In this paper we have analyzed a PGM-based MOEDA named \algNameLong~(\algName) in the context of multi and many objective combinatorial optimization. The main issues investigated in this paper concern a fitness landscape analysis of general-purpose problem features and the analysis of the final PGM structure obtained by \algName. 

We have extracted some features enumerating MNK-landscape problem instances for 2, 3, 5 and 8 objectives.  We aimed to explore the correlation between the problem features and the estimated runtime for \algName~in comparison with NSGA-III, a state-of-the-art algorithm applied to multi and many optimization. In addition, we have evaluated \algName~through the analysis of the final achieved PGM in order to explore one of the main advantages of using EDAs: the possibility of scrutinizing its probabilistic model.

Based on experiments with the MNK instances addressed in this paper, we can conclude that \algName~has a significantly lower estimated runtime compared to NSGA-III, and as expected the feature that has the most significant impact on the estimated runtime is the ruggedness. 
We observed nervetheless that there are features with a negative impact on the efficiency of the model (acting like noisy features).

Furthermore, examining the Pareto front according to a probabilistic view based on the PGM structures, enables the analysis of how the BN can guide the search through specific regions of the Pareto front, being useful to understand the convergence and diversity performance of a given approach.

In the future, other relevant features can be investigated, as those associated with statistics of the probability distribution of points on the Pareto Front.
The approaches will be investigated considering more than eight objectives and other problems.  Additionally, another interesting research direction is the application of other types of PGM that can learn and explore dependencies between variables and objectives.

\section*{Acknowledgment}

The authors would like to express their gratefulness and appreciation to Dr. Arnaud Liefooghe for his insightful feedback and assistance.
M. Delgado acknowledges CNPq grant 309197/2014-7. R. Santana acknowledges support from the IT-609-13 program (Basque Government) and TIN2016-78365-R (Spanish Ministry of Economy, Industry and Competitiveness). 
\bibliographystyle{IEEEtranN}
\bibliography{IEEEabrv,references}

\end{document}